\documentclass[10pt,twocolumn,letterpaper]{article}

\usepackage{cvpr}              
\usepackage{graphicx}
\usepackage{amsmath}
\usepackage{amssymb}
\usepackage{booktabs}
\usepackage{makecell}

\usepackage[pagebackref,breaklinks,colorlinks]{hyperref}

\usepackage[capitalize]{cleveref}
\crefname{section}{Sec.}{Secs.}
\Crefname{section}{Section}{Sections}
\Crefname{table}{Table}{Tables}
\crefname{table}{Tab.}{Tabs.}
\graphicspath{{./figures/}}

\def\cvprPaperID{2537} 
\def\confName{CVPR}
\def\confYear{2023}

\begin{document}

\title{End-to-end Face-swapping via Adaptive Latent Representation Learning}

\author{Chenhao Lin\\
Xi'an Jiaotong University\\
{\tt\small linchenhao@xjtu.edu.cn}
\and
Pengbin Hu\\
Xi'an Jiaotong University\\
{\tt\small hupb666@stu.xjtu.edu.cn}
\and
Chao Shen\\
Xi'an Jiaotong University\\
{\tt\small chaoshen@mail.xjtu.edu.cn}
\and
Qian Li\\
Xi'an Jiaotong University\\
{\tt\small qianlix@xjtu.edu.cn}
}

\twocolumn[{
\renewcommand\twocolumn[1][]{#1}
\maketitle
\begin{center}
    \captionsetup{type=figure}
    \includegraphics[width=.97\textwidth]{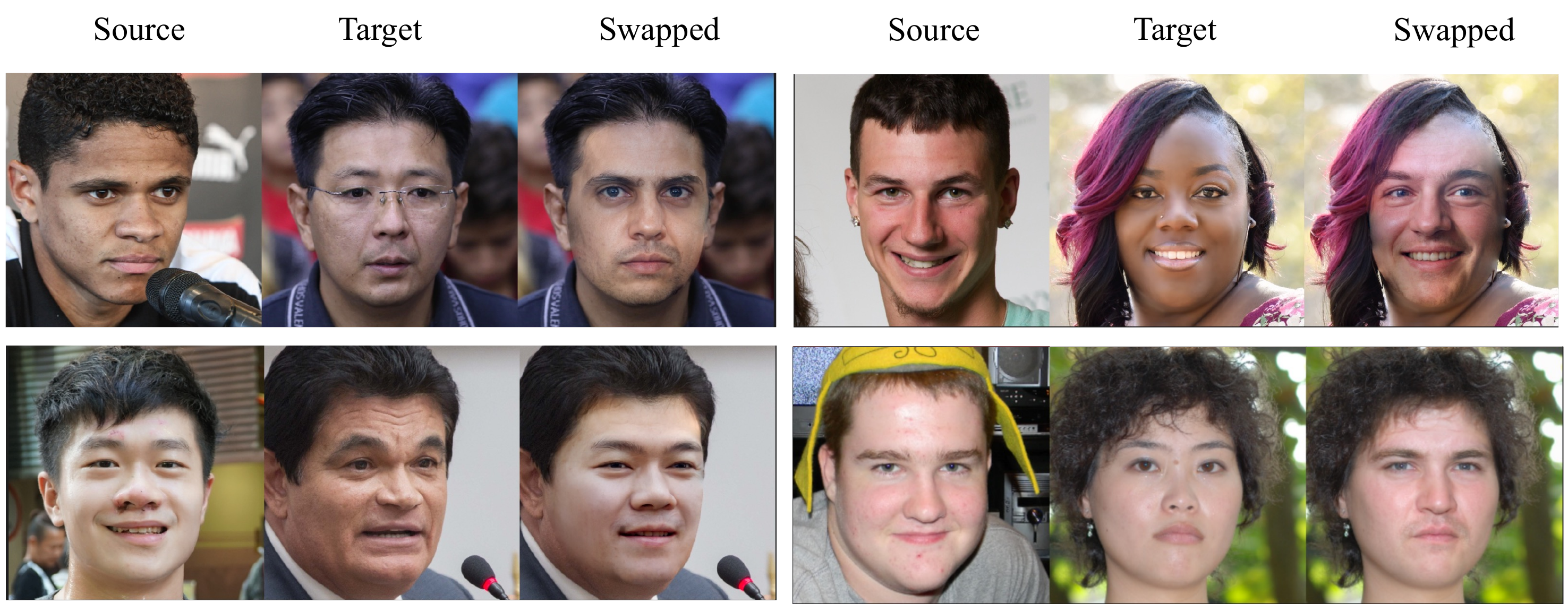} 
    \captionof{figure}{From left to right are source faces, target faces, and swapped faces using our framework.}
    \label{figure1}
\end{center}
}]

\begin{abstract}
Taking full advantage of the excellent performance of StyleGAN, style transfer-based face swapping methods have been extensively investigated recently.
However, these studies require separate face segmentation and blending modules for successful face swapping, and the fixed selection of the manipulated latent code in these works is reckless, thus degrading face swapping quality, generalizability, and practicability.
This paper proposes a novel and end-to-end integrated framework for high resolution and attribute preservation face swapping via Adaptive Latent Representation Learning.  
Specifically, we first design a multi-task dual-space face encoder by sharing the underlying feature extraction network to simultaneously complete the facial region perception and face encoding.
This encoder enables us to control the face pose and attribute individually, thus enhancing the face swapping quality.
Next, we propose an adaptive latent codes swapping module to adaptively learn the mapping between the facial attributes and the latent codes and select effective latent codes for improved retention of facial attributes.
Finally, the initial face swapping image generated by StyleGAN2 is blended with the facial region mask generated by our encoder to address the background blur problem. 
Our framework integrating facial perceiving and blending into the end-to-end training and testing process can achieve high realistic face-swapping on wild faces without segmentation masks. 
Experimental results demonstrate the superior performance of our approach over state-of-the-art methods. \textit{Code will be made publicly available.} 
\end{abstract}

\section{Introduction}
\label{sec:intro}

Face swapping, one of the most popular DeepFake techniques, is receiving growing attention over the last few years. Due to its advantages of the automatic creation process and low cost, face swapping has been extensively used in many fields like film and art creation, education, entertainment, and social media \cite{zhu2021barbershop}. While the malicious use of such techniques also raises serious safety, privacy, and ethical concerns \cite{naruniec2020high,chesney2019deep}.

Face swapping can be defined as the seamless transfer of identity information from the source face image to the target face image while keeping the facial attributes such as expression, skin tone, and lighting unchanged, and producing a fake-like result. 
Early research \cite{Faceswap,perov2020deepfacelab} adopt autoencoder-based frameworks trained on subject-specific images for high-quality face swapping. These methods require a large number of training samples from the same source and target pair, leading to poor generalizability of the models on unseen target faces. To alleviate this limitation, GAN-based subject agnostic face swapping schemes \cite{li2019faceshifter,nirkin2019fsgan} have been proposed. These algorithms generally use public datasets consisting of large enough face samples to train a face swapping model with high generalizability. 
However, the generated deepfake samples usually have low visual quality, like poor identity consistency and attribute preservation, and fail to meet large-scale usage requirements. Therefore, how to simultaneously achieve highly generalizable and realistic face swapping is still a challenging direction to tackle.             

Another challenge of face swapping is maintaining a sense of naturalness while exchanging the contents of two faces and preserving the target image background.  
To achieve this goal, several face swapping methods \cite{korshunova2017fast,dolhansky2019deepfake,jiang2020deeperforensics} learn the commonality in faces and complete face swapping through an encoder-decoder structure without specifying or limiting the location of the area to be swapped, thus resulting in noticeable artifacts of the swapped faces and resolution mismatch in the face regions. Recent studies propose to specify the swap region using face mask images for higher quality face swapping \cite{zhu2021barbershop,nirkin2019fsgan,nirkin2018face}. 
Unfortunately, most swap algorithms require pre-generated segmentation masks or a separately trained segmentation network with a blending module to extract masks and complete face swapping. 
These multi-stage training schemes are complex and may have high computational complexity.  
Moreover, they fail to take advantage of the segmentation mask learning process, which may benefit face swapping. 

More recently, several works \cite{xu2022high,zhu2021one,abdal2021styleflow} adopt a face encoder to map face images into latent codes and then manipulate these codes to get the source to target latent codes. Although such a face encoding mechanism helps to learn the face representation in latent space, the selection and swapping of the manipulated latent codes are fixed and reckless, leading to low-quality swapping of some facial attributes.

To address the limitations, we propose an end-to-end integrated framework for high-resolution Face Swapping via Adaptive Latent representation Learning (\textit{FS-ALL}). Our framework is subject-agnostic and can be applied under different subject faces without additional target-specific training while generating high-quality face swapping images.
Specifically, ALL consists of two main components, Multi-task Dual-space Encoder (MDE) and Adaptive Latent codes Swapping module (ALS). The former is designed to perceive the face swapping region and generate the segmentation mask while simultaneously mapping and decoupling the face image into the face pose and attribute latent codes. Benefiting from the dual-space encoder mechanism and the guidance from the facial perceiving, the proposed MDE can produce a more robust face representation and enhance the face-swapping quality.
ALS introduces a learnable network to measure the latent codes at each layer, enabling us to adaptively select and swap the effective latent codes to obtain the fused latent codes for face swapping with a transformer-based facial attribute retainer.
Then, Stylegan2 is selected as a decoder to obtain the preliminary face-swapped image. Finally, this image and the corresponding mask region are passed through an internal facial blending module to complete the face swapping. Our contributions can be summarized as follows,
\begin{itemize}
\item We propose a novel end-to-end integrated framework, which can elegantly produce high-resolution and attribute preservation face-swapping via adaptive latent representation learning (ALL). 
\item We carefully design ALL to perform face-to-latent space mapping and decoupling and perceive facial regions simultaneously for robust face representation learning. ALL also helps to adaptively select and swap the effective latent codes for face swapping with enhanced attribute preservation.
\item Experimental results demonstrate that the face-swapped images generated using our framework are improved in terms of naked eye effect and quantitative metrics compared with the baselines and state-of-the-art face swapping methods.
\end{itemize}

\begin{figure*}[ht!]
\centering
\includegraphics[width=0.95\linewidth]{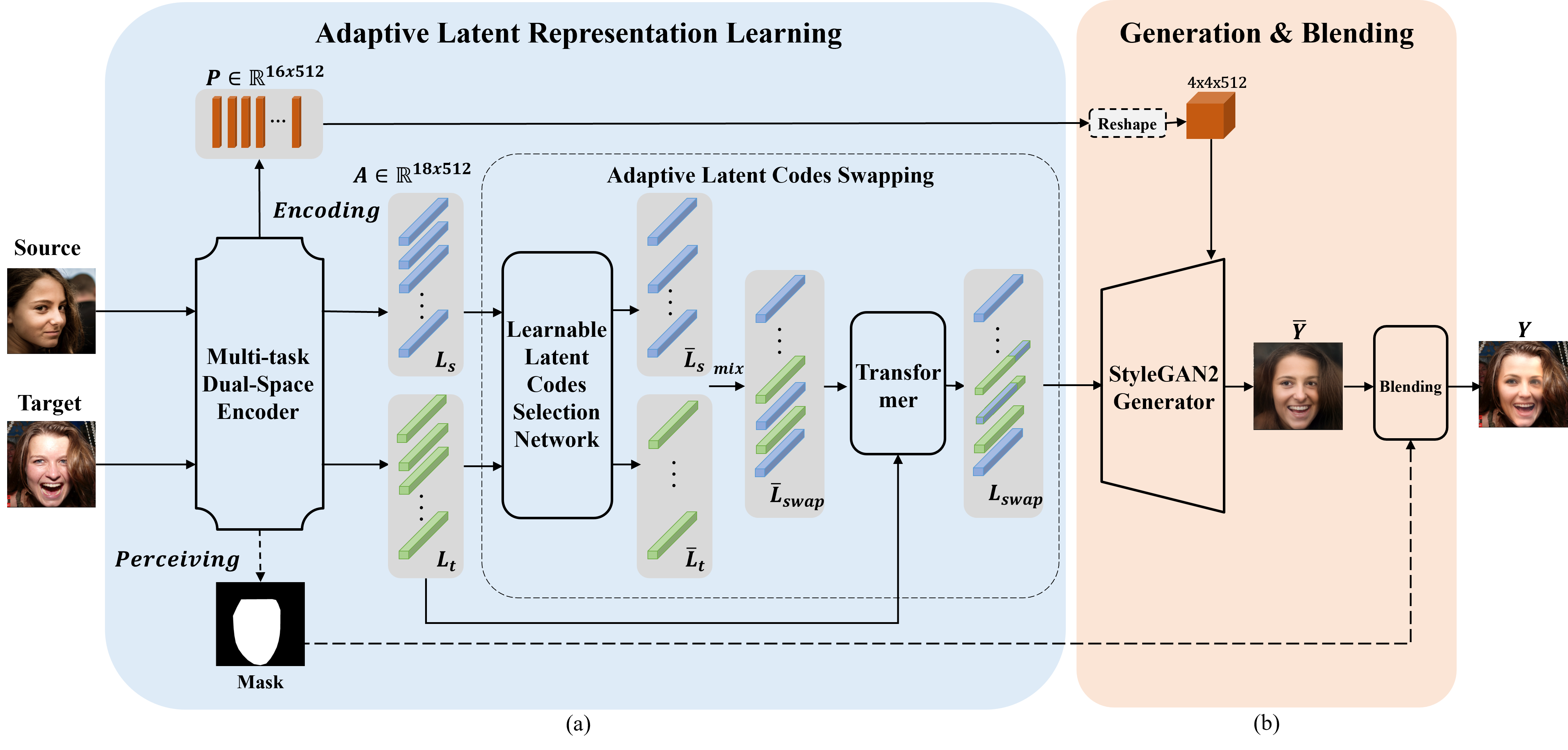}
\caption{Overview of the proposed framework. (a) ALL includes a multi-task dual-space encoder to map the source and target face pairs to $P$ and $A$ latent spaces, and achieve target facial region perceiving and segmentation mask generation. An adaptive latent codes swapping module feeds the source $L_s$ and target $L_t$ latent codes into a learnable network and obtains $\bar{L}_{swap}$. The latent codes of $\bar{L}_{swap}$ and the target ${L}_{t}$ are fed into the transformer to get $L_{swap}$. (b) G\&B feeds $L_{swap}$ into the generator to get the preliminary swapped face image $\bar{Y}$, and blends it with the mask to get the swapped face $Y$. The solid line represents the encoding task and dash line perceiving task. 
\label{overview}}
\end{figure*}

\section{Related Works}
\label{sec:related}
\subsection{Autoencoder methods}Face swapping algorithms were first implemented by an encoder and a set of decoders. The faces were encoded and fed into each other's encoders to achieve face swapping, e.g., Deepfake\cite{Deepfake}. In \cite{korshunova2017fast}, the authors proposed to improve face swapping by preserving face pose, expression, and lighting using CNN. These approaches usually require many person-specific images or video data to support the model's training, limiting their practicality.

\subsection{3D face methods}
Several face swapping algorithms are based on the 3D face model\cite{lyu2022portrait, otberdout2022sparse}. Face2Face\cite{thies2016face2face} transferred expressions from source to target face by fitting a 3D morphable face model (3DMM)\cite{thies2015real}. Nirkin\cite{nirkin2018face} used a fixed 3D face shape as the proxy. The expression of the target face is copied, and then the face is segmented by a separately trained FCN\cite{long2015fully} to generate the corresponding mask and transfer the face. Most of the methods usually perform face reproduction first, then cut and paste the face into the target using an additional segmentation network.

\subsection{GAN-based methods}
FSGAN\cite{nirkin2019fsgan} is the current face replacement algorithm that achieves one of the SOTA results. Four GANs are trained separately to accomplish face reproduction, face segmentation, face inpainting, and face fusion. The reproduced face is obtained by differencing the Euler angles between the source face and the target face, then segmenting, cropping, and fusing them. IPGAN\cite{bao2018towards} and FaceShifter\cite{li2019faceshifter} divided faces into identity and attribute information, and FaceShifter inserted the identity information into the attribute information continuously and completed the training by the GAN network.

\subsection{Latent space manipulation}
Recently, several studies using the pre-trained StyleGAN\cite{xu2022high, zhou2022pro, xu2022region, sun2022fenerf, he2022gcfsr} have proposed to manipulate the latent codes of images for the purpose of facial content manipulation. 
The GAN Inversion found the latent code of the image and operated on it to modify the image.
One is the optimization-based approach of optimizing the latent code to minimize the error for the given image. Barbershop\cite{zhu2021barbershop} combined the latent code and mask to edit the specified region. Transeditor\cite{xu2022transeditor} controlled the face pose and face style individually by mapping the face into the double space and found the average vector between the latent code of the two faces for linear transformation to achieve facial attribute editing. The optimization-based approach is similar to the target-specific face-swapping algorithm, which is more advantageous in generation quality but less efficient.
The other one is to train a generic swap to edit the latent codes. MegaFS\cite{zhu2021one} proposed to transfer the identity from a source image to the target by a non-linear trajectory without explicit feature disentanglement. FSLSD\cite{xu2022high} designed a framework to disentangle the latent semantics of a pretrained StyleGAN to transfer the source identity and preserve target appearance. 
Our approach improves MegaFS and FSLSD in latent representation learning by perceiving facial regions, adaptively selecting and swapping the latent codes, and introducing an internal facial blending module. 

\section{Methodology}

The proposed framework is composed of two major phases, namely the adaptive latent representation learning and the generation\&blending, as illustrated in Figure \ref{overview}. 

In the ALL phase, a feature extraction sharing network with a multi-task learning module is implemented, to perceive the face-swapping regions and map the face into face pose space $P$ and face attribute space $A$. In addition, an adaptive latent codes swapping module is designed to select, swap and fuse the source and target latent codes adaptively. In the G\&B phase, the fused latent codes and the latent codes in $P$ space are fed into the StyleGAN2 generator \cite{karras2020analyzing} and an internal blending module together with the segmentation masks to obtain the swapped face.

\begin{figure}[!t]
\centering
\includegraphics[width=0.95\linewidth]{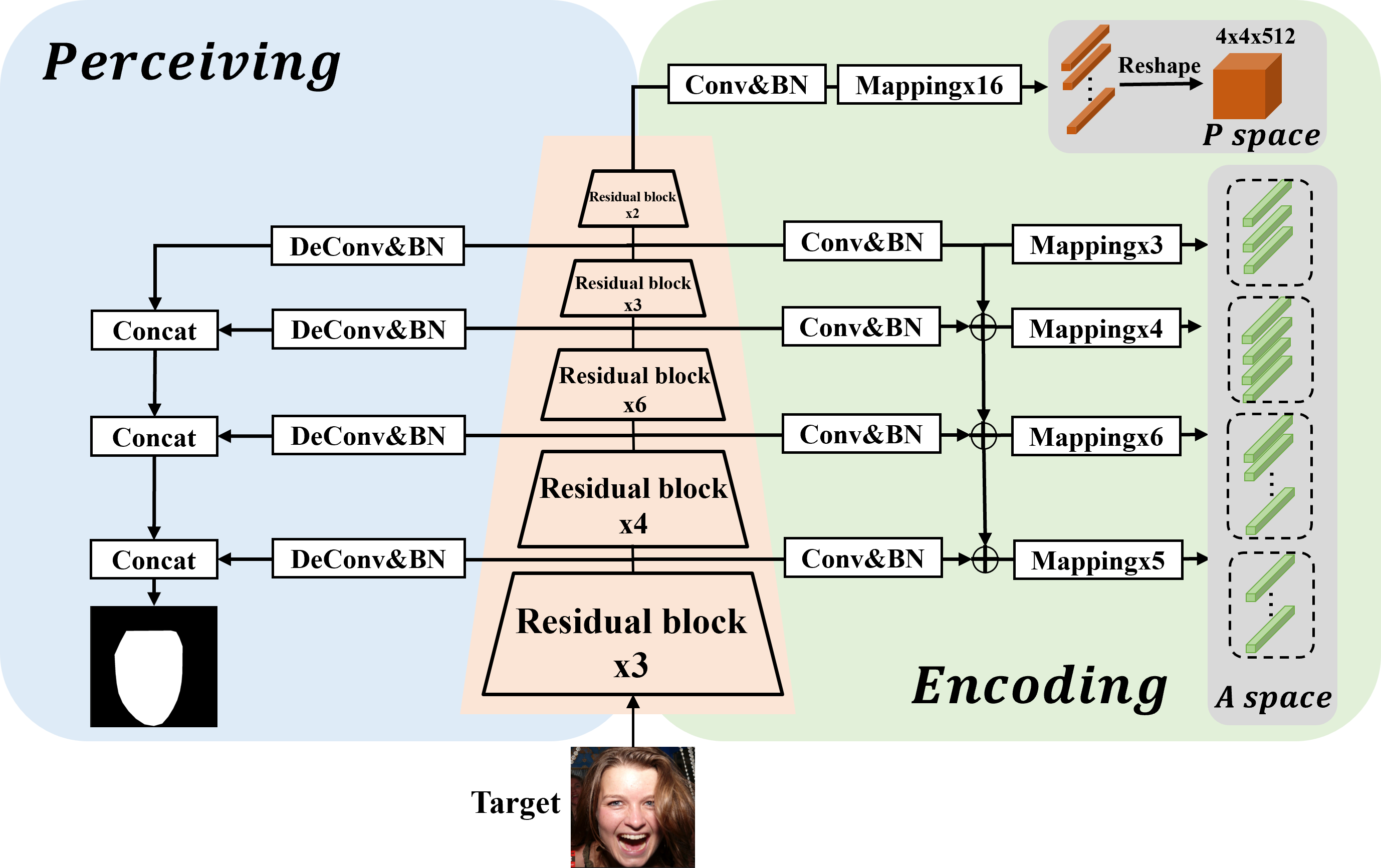}
\caption{Multi-task Dual-Space Encoder architecture. ResNet 50 is used as a feature extraction network, and the left-side facial perceiving task uses U-Net for dimensional connectivity. The right-side encoding task maps the faces into $P$ space and $A$ space based on the FPN.
\label{Encoder}}
\end{figure}
\subsection{Adaptive latent representation learning} 
As shown in Figure \ref{overview}, ALL mainly consists of two components, Multi-task Dual-space Encoder (MDE) and Adaptive Latent codes Swapping module (ALS).

\subsubsection{Multi-task dual-space encoder}
To achieve simultaneous face swapping-region perception and face-to-latent space mapping and decoupling, MDE first applies ResNet\cite{he2016deep} as the shared backbone for multi-scale feature extraction from the input face image, as illustrated in Figure ~\ref{Encoder}. The top layers of the network are split into two branches and the first is for \textit{face perceiving}. The multi-scale feature maps are concatenated by channel dimension and passed through the upsampling, batchnorm, and leakyReLU layers with reference to the U-Net structure to generate the segmentation masks. 

The second branch is for \textit{face encoding}. We introduce dual-space face inversion to map the faces into two separate latent spaces, pose latent space $P$, and attribute latent space $A$, expecting to control the face pose and facial attributes such as expression, skin tone, identity information, and hairstyle, separately.  
Specifically, the feature maps at each scale are fed into the FPN and mapped to obtain the latent code $A \in \mathbb{R}^{18\times512}$ in $A$ space, representing facial attributes and identity information. We expand the number of scales of the feature maps to make the network pay attention to the details of the face image and find it helps to generate higher face quality reconstruction. The latent code in $P$ space is passed through the feature map output from the top layer of the encoder and then through 16 non-linear mapping networks to obtain the latent code $P \in \mathbb{R}^{16\times512}$ in $P$ space. $P$ is used as the base input of the pre-trained StyleGAN2 generator, which is reshaped into $P \in \mathbb{R}^{4\times4\times512}$ as the input of the generator.

Thus, the proposed MDE can accurately perceive the facial regions for end-to-end face-swapping and simultaneously provide guidance to face encoding. It also maps the face into the $P$ space and $A$ space to transfer the attribute and identity information of the source face to the target face in the subsequent exchange of latent codes while keeping the face pose unchanged.


\subsubsection{Adaptive latent codes swapping}
The selection of latent codes in existing approaches such as pSp\cite{richardson2021encoding}, MegaFS\cite{zhu2021one} and FSLSD\cite{xu2022high} is usually fixed and reckless. Thus, some facial attributes that are represented by accumulating specific multiple latent codes may not be well learned and swapped, resulting in poor attribute preservation. To alleviate this limitation, we propose a Learnable Latent codes Selection Network (L2SNet) along with a transformer-based attribute retainer to adaptively select and swap the latent codes while keeping the facial attributes.

\begin{figure}[]
\centering
\includegraphics[width=0.9\linewidth]{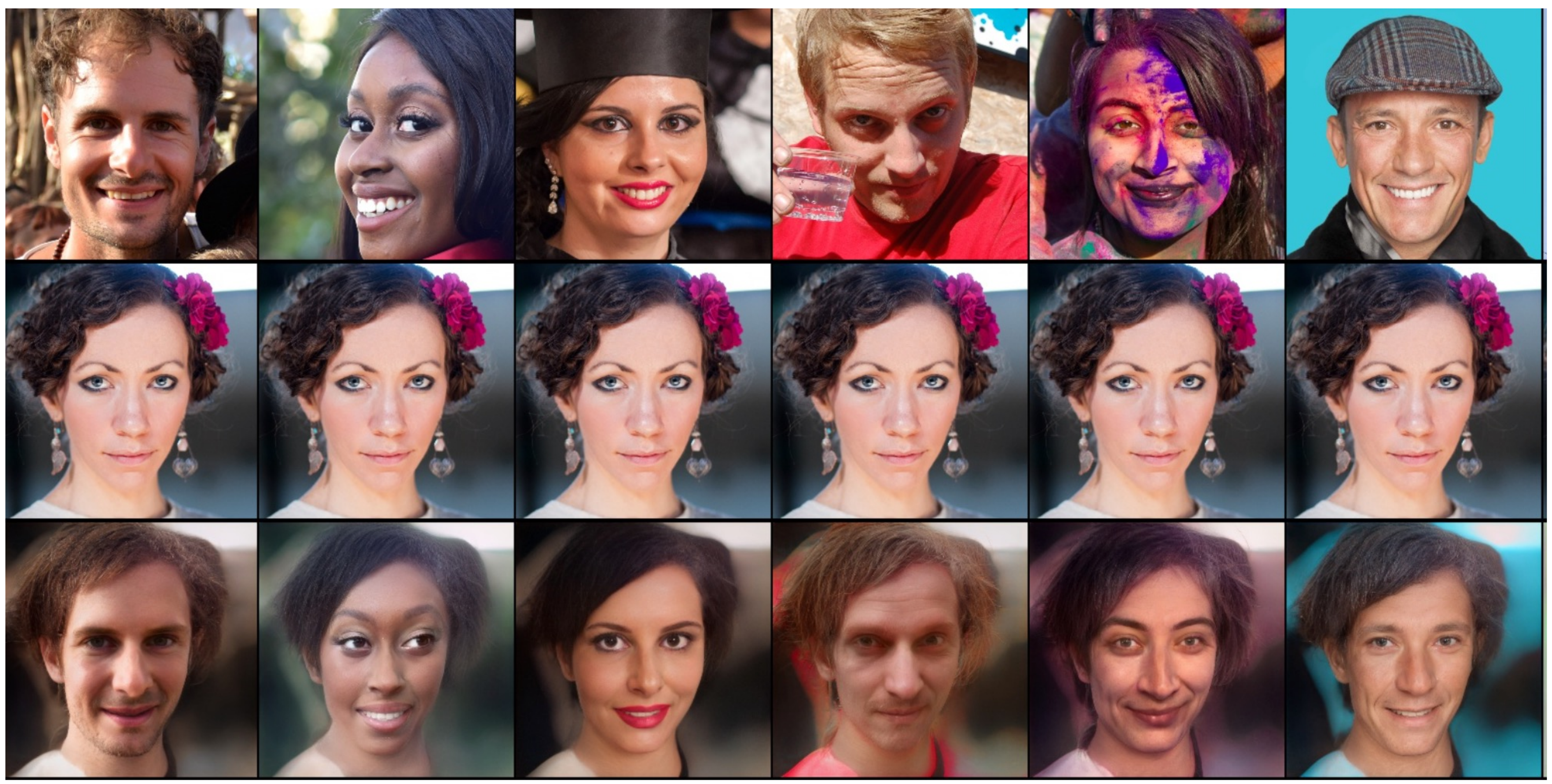}
\caption{Row 1 and row 2 show the source and target faces. By keeping the first three latent codes of the targets in $A$ space and all latent codes in $P$ space, we can make the source have the same pose as the target.
\label{fixpose}}
\vspace{-0.2cm}
\end{figure}

\begin{figure}[]
\centering
\includegraphics[width=0.95\linewidth]{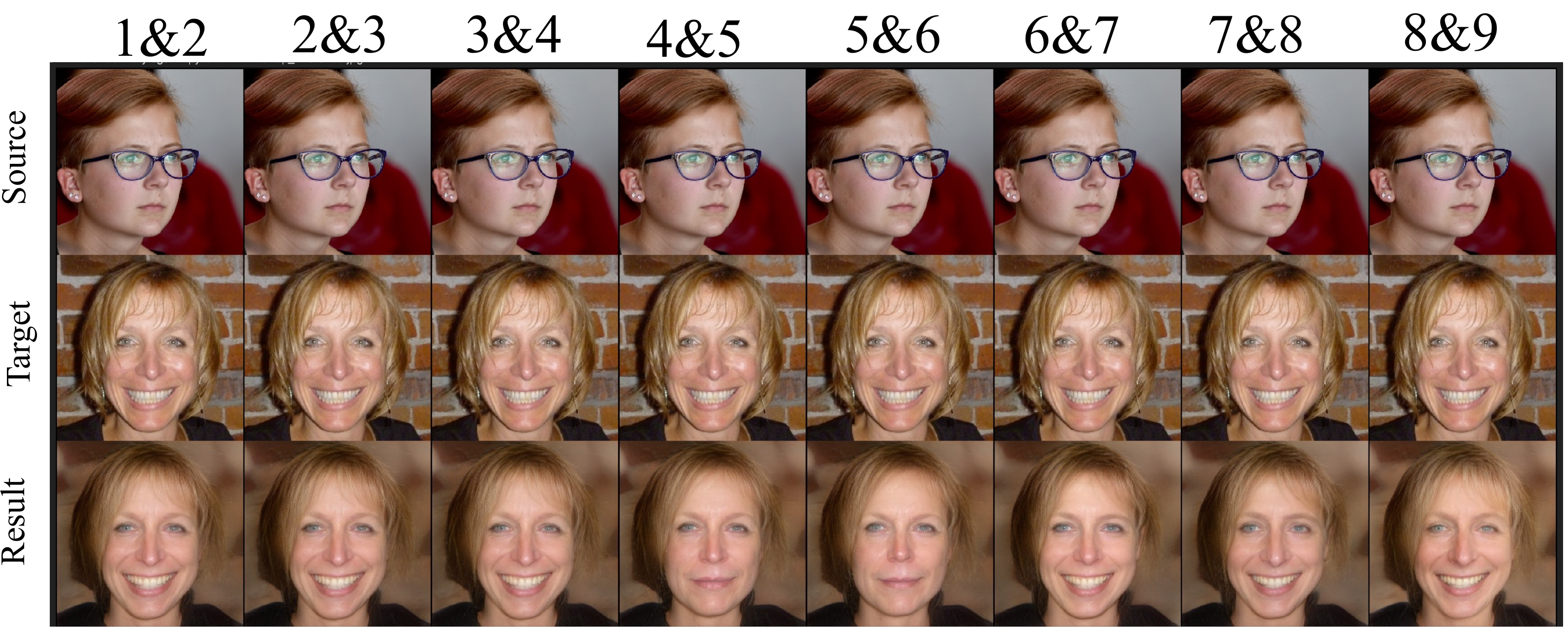}
\caption{By changing the latent codes in $A$ space two by two (e.g., 1\&2, 2\&3 latent codes), we find that the fifth latent code affects the information around the mouth.
\label{mouth}}
\vspace{-0.1cm}
\end{figure}

\textbf{Learnable latent codes selection network.}
Inspired by the findings in pSp\cite{richardson2021encoding}, we assume that different latent codes may correspond to different facial attributes. 
Then we conduct a preliminary experiment and find that all the latent codes in $P$ space and the first three latent codes in $A$ space are used to control the face pose, as shown in Figure \ref{fixpose}. The other latent codes in $A$ space are mainly used to control the non-pose attributes. For example, the fifth latent code mainly affects the information around the face's mouth, as shown in Figure \ref{mouth}. 
Therefore, we propose to fix all the latent codes in $P$ space and the first three latent codes in $A$, and to adaptively learn and select effective latent codes in the remaining codes for face swapping. In SENet\cite{hu2018squeeze}, the claim that the weight of each channel of the image is different inspires us that the content of each latent code of the face image may also be different. To this end, we design a learnable network L2SNet with SENet structure to measure the score of each latent code of the source and target faces separately as follows.
\begin{equation}
\begin{aligned}
Score = L2SNet(L)
\end{aligned}
\end{equation}
\begin{equation}
Score_{st} = concat(score_{s},score_{t}) = 
\begin{bmatrix}
S_{4} & T_{4}\\
S_{5} & T_{5}\\
\vdots & \vdots\\
S_{18} & T_{18}\\
\end{bmatrix}
\end{equation}
where $Score$ is the output of each latent code $L$ of $L2SNet$. $s$ and $t$ represent the source and target faces. $Score_{st}$ represents the group of latent codes after concatenation, and $S_{i}$, $T_{i}$ represents the $i^{th}$ latent code's score from top to bottom, respectively. Then we calculate $Mask_{L}$,
\begin{equation}
Mask_{L} = argmax(Score_{st}^{1},Score_{st}^{2}\cdots\cdots,Score_{st}^{18})
\end{equation}
where $argmax(\dots)$ outputs a set of one-hot codes to select the effective latent codes.

For the training of L2SNet, we need to obtain the gradient of the weights of the evaluation network. Unfortunately, the $argmax$ function is not differentiable. To solve this problem, we use the approximate gradient \cite{chen2021dynamic} as the weight gradient of the L2SNet, as shown in the following:
\begin{equation}
\bar{{score}_{st}^{i}}= \frac{e^{score_{st}^{i}}}{\sum_{n} e^{score_{st}^{i}}} \ \ \ \ \ \ \ \  i \in 
\begin{bmatrix}
0 & m-1
\label{softmax}
\end{bmatrix}
\end{equation}
In the backward propagation process, we adopt the $softmax$ function Eq.$(\ref{softmax})$, to get $\bar{{score}}_{st}^{i}$ as close to 0 or 1 as possible to shorten the gap between the generated mask and the unique thermal code. And $softmax$ is differentiable, making generated $Mask_{L}$ continuously optimized.

\textbf{Latent codes swapping with transformer-based attribute retainer.}
Then we swap the selected latent codes with a transformer to achieve enhanced attribute preservation.
$L_{st}$ dot product with $Mask_{L}$ to get the preliminary selected latent code $\bar{L}_{swap}$
\begin{equation}
\bar{L}_{swap} = L_{st} \cdot Mask_{L}
\end{equation}
Our preliminary experiment also observed that some facial attributes are mixed together in several different latent codes. Thus, the attribute cannot be accurately controlled if we swap a single latent code. While if several latent codes are swapped together, the other attributed will be impacted. Therefore, we design a transformer-based attribute retainer to adaptively decouple the facial attributes in different latent codes and re-mix them for more precise control of the attributes.  
In our transformer, the latent codes of the target, $L_{t}$ are used as query ($Q$), i.e., as the attention query statement. The swapped latent codes $L_{swap}$ are used as key ($K$) and value ($V$). It can be represented as follows,
\begin{equation}
Q = \bar{L}^{i}_{t}W^{Q},K = \bar{L}^{i}_{swap}W^K,V = \bar{L}^{i}_{swap}W^V
\end{equation}
\begin{equation}
L^{i}_{swap} = softmax(\frac{QK^T}{\sqrt{d_{k}}})V + \bar{L}^{i}_{swap}
\end{equation}
where $W^{Q}, W^{K}, W^{V}$ are linear projection matrices, and $dk$ is the dimensionality of the latent code. $L_{swap}$ receives queries from $L_{t}$. 

\subsection{Generation and blending }
The pre-trained model of StyleGAN2 is used as a generator, the target latent code $L_p$ in $P$ space is used as the base input, and ${L_{swap}}$ is fed as the style to obtain the preliminary face swapping image $\bar{Y}$.
\begin{equation}
\bar{Y} = G(L_p,L_{swap})
\end{equation}
To address the background vignetting, hair color hairstyle with certain artifacts, and other problems, we design an internal blending module to seamlessly connect the $Y$ facial region to the target based on Poisson fusion\cite{perez2003poisson}. We first soften the area around the face mask so that the face is smoothly connected to the surrounding area. Then we use the softened mask to separate the face area and fuse it with the target.
Existing methods usually apply the blending module as image post-processing. In comparison, we propose an elegant way of incorporating Poisson fusion into our framework to achieve end-to-end training and testing by solving the problem of gradient non-passing during the backward propagation process.

\subsection{Training losses}
The overall training loss of the proposed method consists of facial perception and encoding loss $\mathcal {L}_{MDE}$ and latent code swapping loss $\mathcal {L}_{ALS}$. 
\begin{equation}
 \mathcal {L} =\mathcal {L}_{MDE}+\mathcal {L}_{ALS}
\end{equation}
\noindent\textbf{Training losses of MDE.}   
Since MDE completes the face perception and latent space mapping, its loss function, ${L}_{MDE}$, consists of two parts. We use binary cross-entropy, ${L}_{p}$, as the loss function for the face perception task. Another part of the loss function, ${L}_{inv}$, is responsible for training the mapping of the face to the latent space. 
It includes the reconstruction loss $\mathcal{L}_{rec}$ at the pixel level and id loss $\mathcal{L}_{id}$ to calculate the loss of identity information between the reconstructed face and the real face using Arcface\cite{deng2019arcface}. $\mathcal{L}_{ldm}$ calculates the landmark loss to keep the face pose stable using the facial landmark extractor\cite{king2009dlib}. $\mathcal{L}_{LPISP}$ measures the LPISP loss\cite{zhang2018unreasonable} and finally passes the reconstructed face through the MDE again to calculate the $\mathcal{L}_{latent}$ loss at the latent code level.
The overall loss function is as follows:
\begin{equation}
 \mathcal {L}_{inv} =\lambda_{1}\mathcal {L}_{rec}+\lambda_{2}\mathcal {L}_{id}+\lambda_{3}\mathcal {L}_{ldm}+\lambda_{4}\mathcal {L}_{latent}+\lambda_{5}\mathcal {L}_{LPIPS}
\end{equation}
\begin{equation}
\mathcal {L}_{MDE} = \psi\mathcal {L}_{p} +\mathcal {L}_{inv}
\end{equation}
where $\psi,\lambda_{1},\lambda_{2},\lambda_{3},\lambda_{4}$ and $\lambda_{5}$ are loss weights.

\noindent\textbf{Training losses of ALS.}
For the training of adaptive latent representation learning, the reconstruction loss, id loss, landmark loss, and LPIP loss are still included, with slight differences in the objects from the previous ones. Please refer to the supplementary material for all specific loss functions. The overall loss function is as follows:
\begin{equation}
 \mathcal {L}_{ALS} =\gamma_{1}\mathcal {L}_{rec}+\gamma_{2}\mathcal {L}_{id}+\gamma_{3}\mathcal {L}_{ldm}+\gamma_{4}\mathcal {L}_{LPIPS}
\end{equation}
where $\gamma{1},\gamma{2},\gamma{3}$ and $\gamma{4}$ are loss weights. Here $y$ represents the final face replacement image after blending.

\section{Experiments}
In this section, two groups of comparison experiments with several state-of-the-art methods are performed to validate the effectiveness of the proposed approach. We first compare FSALL with the same type of SOTA methods, MegaFS\cite{zhu2021one} and FSLSD\cite{xu2022high}, on high-resolution datasets, CelebA-HQ\cite{karras2017progressive} and FFHQ\cite{karras2019style}. Then the current mainstream face swapping algorithms DeepFake\cite{Deepfake}, FaceSwap\cite{Faceswap}, and FaceShifter\cite{li2019faceshifter} are performed on FaceForensics++\cite{rossler2019faceforensics++} for comparison. Both qualitative and quantitative comparisons are performed in two groups of experiments. All experiments were conducted in Pytorch on two Tesla A100 GPUs and two Intel XEON Gold CPUs.

Following previous SOTA methods \cite{li2019faceshifter,zhu2021one,xu2022high}, we use several metrics, including ID retrieval, ID similarity, pose errors, and expression errors, for quantitative evaluation of face swapping. Failure rate and Frechet Inception Distance (FID) are adopted to measure the quality of reconstructed faces. \textcolor{black}{Details are described in the supplementary material.} 

\subsection{Experiments on high-resolution datasets}

\begin{figure*}[ht!]
\centering
\includegraphics[width=0.6\linewidth]{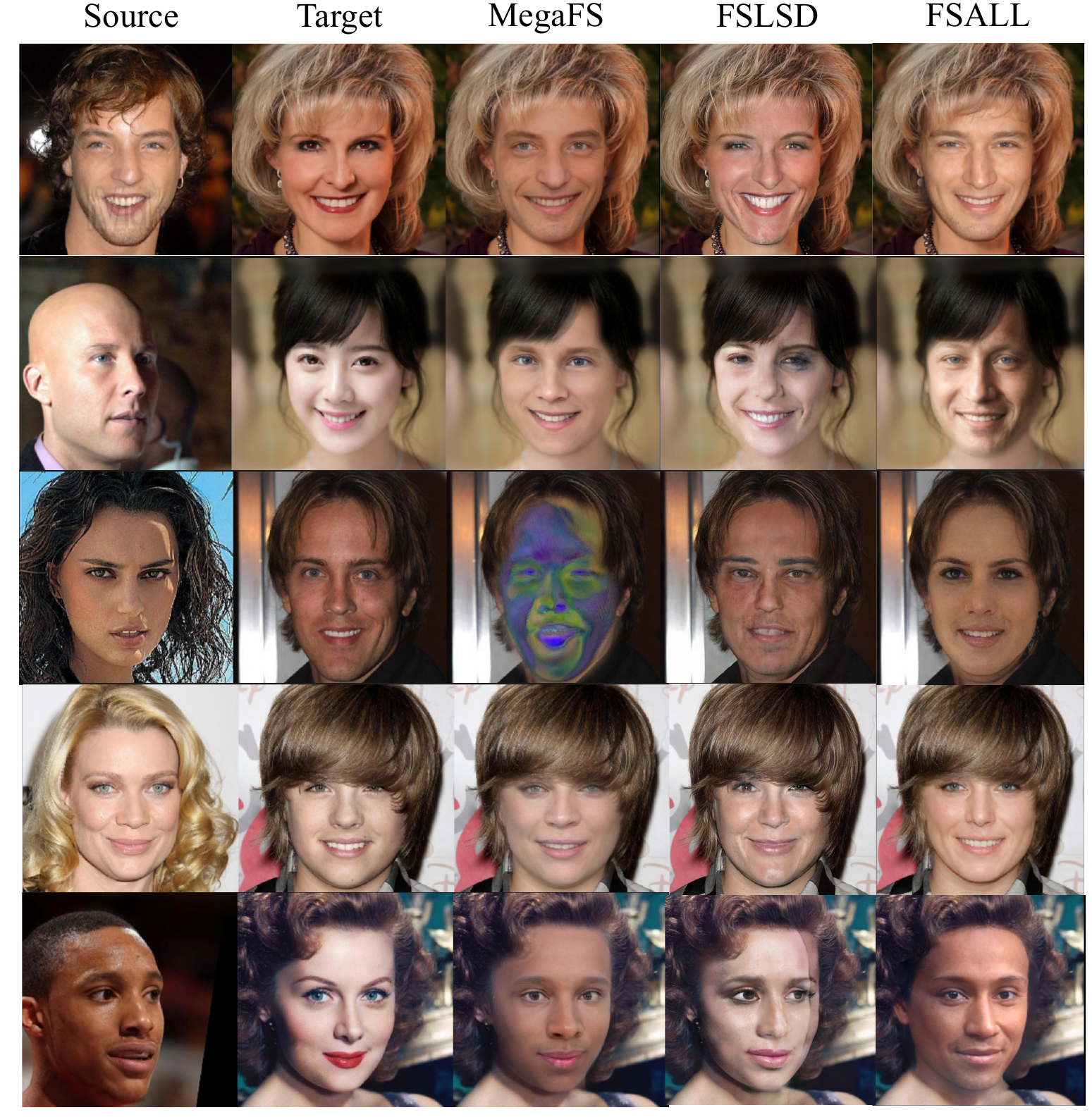}
\caption{Qualitative comparison of face swapping on CelebA-HQ dataset. We can see that FSALL can produce higher quality swapped faces with more source identity information while maintaining more of the target face attributes such as mouth region and eye region.
\label{CelebA}}
\vspace{-0.2cm}
\end{figure*}

\noindent\textbf{Qualitative comparison on CelebA-HQ.}
Since both MegaFS and FSLSD methods require pre-segmented facial masks for face swapping, we first conducted experiments on CelebA-HQ with additional CelebAMask-HQ. While our method only uses face images during test.

The qualitative comparisons are shown in Figure \ref{CelebA}. It can be seen that FSALL has more advantages in terms of facial attribute control. We attribute the success to our ALS module, which adaptively manipulates the effective latent codes, resulting in better identity transfer and attribute preservation. In comparison, FSLSD and MegaFS operate on the fixed first seven latent codes for each face change operation, which is difficult to guarantee that it will work for every face. As in rows 2 and 4, the attributes around the mouth of the FSLSD do not keep well with the target face, and the eye attributes of the MegaFS in rows 1 and 5 do not keep well with the target face either. Our method also shows a natural-looking swapping and a good performance on the identity information transfer of source faces due to the robust face representation learned by FSALL.

In addition, we can see the swapped faces generated by FSLSD tend to be blurred (rows 2,3,5). Although FSLSD designed a target encoder and decoder module to achieve low-to-high-resolution facial image restoration and reduce the artifacts around the face, it may also result in blurred internal facial regions. Figure \ref{CelebA} also shows that the swapped faces of MegaFS lack facial details due to the limited representation ability of the encoder.

\begin{table}[!t]
\begin{tabular}{c|c c c}
    Method & ID similarity$\uparrow$ & pose$\downarrow$ & expression$\downarrow$\\
    \hline{}
    MegaFS &0.3781  & 3.72 & 2.92 \\
    FSLSD &0.3980  & \textbf{3.55} & 2.81 \\
    FSALL & \textbf{0.4327} & 3.61 & \textbf{2.70}\\
\end{tabular}
\caption{Quantitative evaluation on CelebA-HQ.}
\label{CelebA_table}
\vspace{-0.4cm}
\end{table}
\noindent\textbf{Quantitative comparison on CelebA-HQ.}
We generated 
swapped faces on CelebA-HQ for quantitative comparison, shown in Table \ref{CelebA_table}. The lower ID similarity and expression error demonstrate that FSALL maintains more source identity information and achieves better attribute control and preservation. FSALL is slightly inferior to FSLSD in pose error because FSLSD adopted an effective but external landmark estimator to align source and target landmarks. Their blending module makes the target to be constantly modified based on the source and helps the pose transfer. However, this module may also result in poor identity transfer and distortion in the facial region.

\begin{figure}[t!]
\centering
\includegraphics[width=0.8\linewidth]{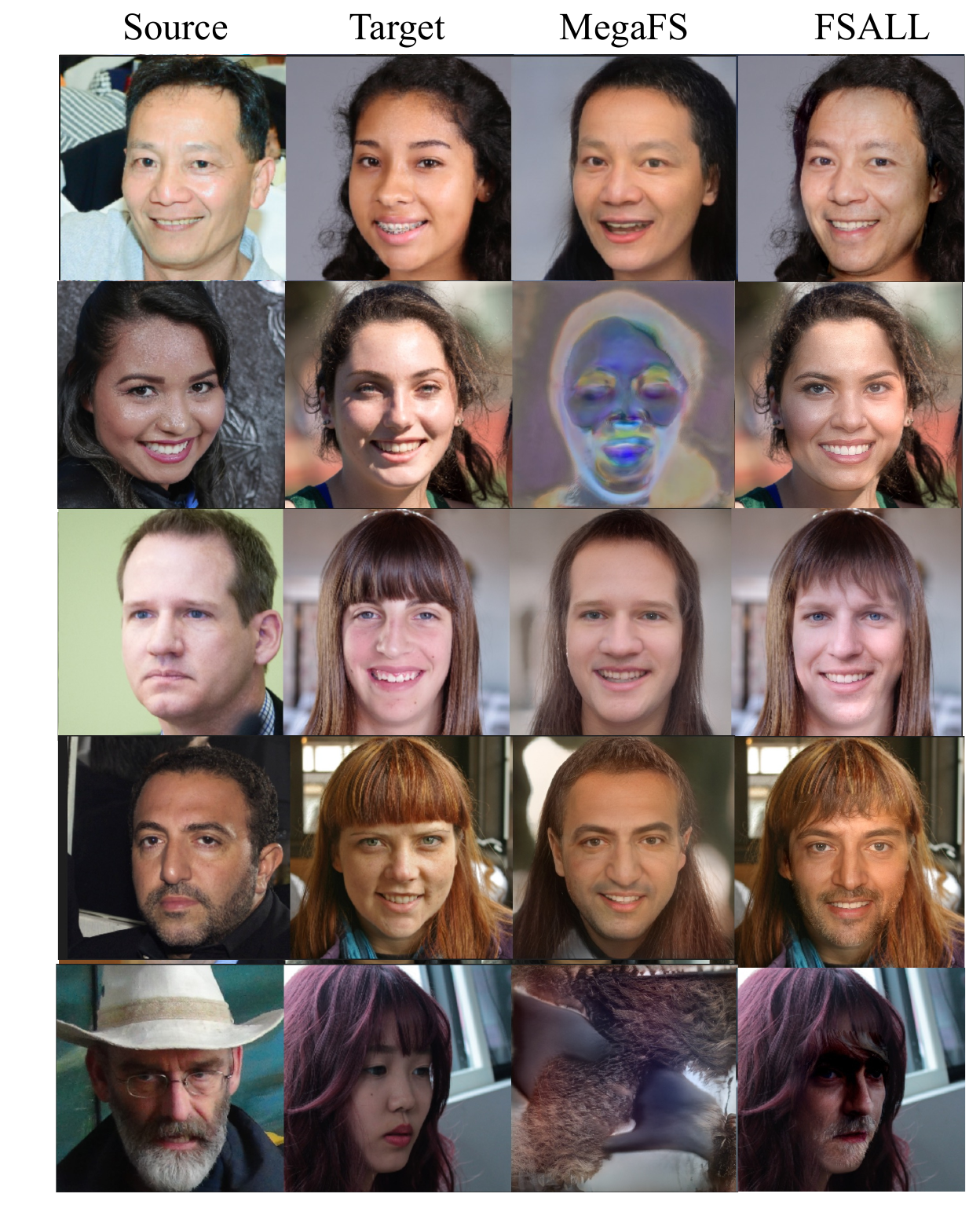}
\caption{Qualitative comparison with MegaFS on FFHQ. FSALL accomplishes more face swapping and retains more detail in the characters, including the challenged face swapping (row 5).
\label{FFHQ}}
\end{figure}
\noindent\textbf{Qualitative comparison on FFHQ.}
We also performed comparisons on FFHQ. As shown in Figure\ref{FFHQ}, FSALL completes all the face swapping tasks, while MegaFS shows some failures (rows 2 and 5), because the MegaFS encoder fails to map the faces into the correct latent spaces. Moreover, the background and hairstyle of MegaFS swapped faces are blurred, differing wildly from the targets (rows 3 and 4). In comparison, FSALL retains them thanks to the proposed facial perceiving and internal blending module. FSALL also retains more detailed facial attributes of the source, such as the beard in rows 4 and 5.

\begin{table}[!t]
\begin{tabular}{c|c c c}
    Method & ID retrieval(\%)$\uparrow$ & pose$\downarrow$ & expression$\downarrow$\\
    \hline{}
    MegaFS &85.93  & 5.83 & 3.05 \\
    FSALL & \textbf{87.61} & \textbf{5.57} & \textbf{2.96}\\
\end{tabular}
\caption{Quantitative evaluation on FFHQ.}
\label{FFHQ_table}
\vspace{-0.4cm}
\end{table}

\noindent\textbf{Quantitative comparison on FFHQ.}
Swapped faces on FFHQ were produced to perform quantitative measurements, including ID retrieval, pose, and expression, following MegaFS. The results are shown in Table \ref{FFHQ_table} and FSALL achieves superior performance in all metrics. FSALL maintains more source identity information and illustrates better attribute preservation due to the multi-task dual-space encoding in MDE and the superiority of ALS.

\subsection{Experiments on FaceForensics++}
We additionally evaluated FSALL on the low-resolution dataset FaceForensics++. It contains many real videos from the physical world of Youtube with varying quality. We applied dlib \cite{king2009dlib} to detect the face in video frames, crop out face rectangle images, and filter out low-quality images. Several popular methods were used for comparison. 

\begin{figure}[t!]
\centering
\includegraphics[width=1\linewidth]{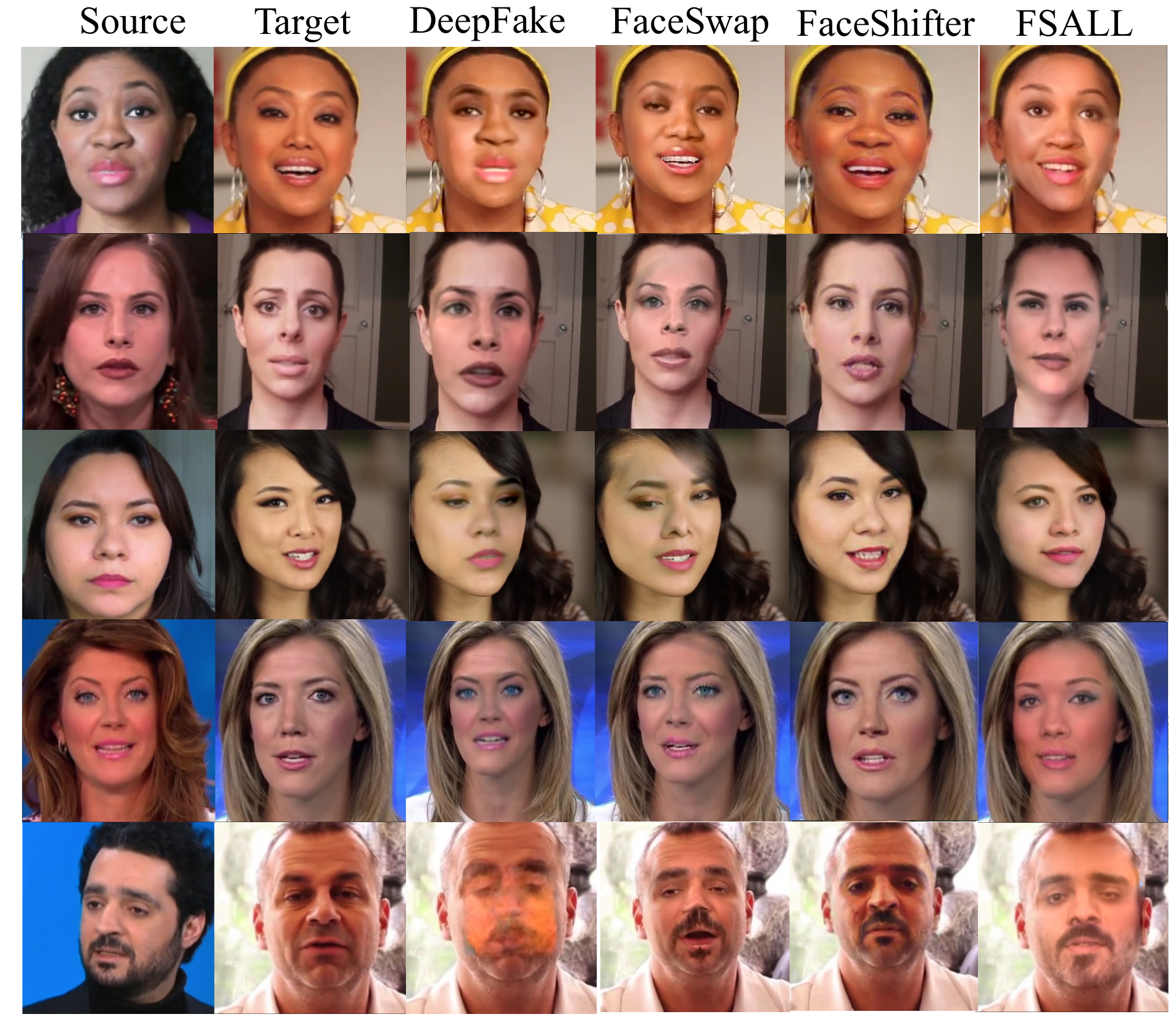}
\caption{Qualitative comparison with DeepFake, FaceSwap, and FaceShifter on FaceForensics++. FSALL passes more source identity information. 
\label{FF++}}
\end{figure}

\noindent\textbf{Qualitative comparison.} Figure \ref{FF++} shows that both DeepFake and FaceSwap have failure cases. In row 3, DeepFake and FaceSwap have errors around the eyes. In the last row, DeepFake directly fails to render the portrait, and FaceSwap has apparent mistakes in the nose and eye areas. Both FaceShifter and FSALL can achieve successful face swapping, and the swapped faces of FaceShifter may be closer to the target faces than FSALL in terms of expression properties (row 3). However, FaceShifter still has incomplete face swapping. For example, FaceShifter barely passes the source's identity information in row 4. In row 2, FaceShifter has a noticeable artifact on the right face.

\noindent\textbf{Quantitative comparison.} 
The results are shown in Table \ref{table_FF++}. FSALL and FaceShifter achieve good results, where FSALL obtains the highest ID retrieval and slightly underperformance in pose and expression error. This is because the swapped latent code still contains part of the character attribute information, which indicates that there is still room for improving the decoupling of the latent code in our algorithm on FaceForensics++.
FaceShifter adds the identity information of the source to the target to keep the same pose and expression. However, it is easy to produce incomplete or no face replacement, such as in row 4 in Figure \ref{FF++}. The higher ID retrieval value also indicates that FSALL has a better identity information transfer.

\begin{table}[!t]
\centering
\begin{tabular}{c|c c c}
    Method & ID retrieval(\%)$\uparrow$ & pose$\downarrow$ & expression$\downarrow$ \\
    \hline{}
    DeepFake & 80.54 & 3.80 & 3.04 \\
    FaceSwap & 67.29 & \textbf{2.31} & 2.80 \\
    FaceShifter & 88.46 & 3.08 & \textbf{2.54} \\
    \hline{}
    FSALL & \textbf{90.23} & 3.10 & 2.84 \\
\end{tabular}
\caption{Quantitative evaluation on FaceForensics++}
\label{table_FF++}
\vspace{-0.5cm}
\end{table}
\subsection{Ablation Study}
We performed ablation experiments to explore the effect of the proposed MDE and ALS components. More ablation studies can be found in the supplementary material. 

To evaluate MDE, we generated 
reconstructed face images using FFHQ and CelebA-HQ datasets and performed quantitative measurements.
The \textit{HieRFE} encoder in MegaFS and DE (without multi-task learning) are used for comparison.
The quantitative results for reconstructed faces are shown in Table \ref{Ab_Encoder_table1}. 
We can see MDE performs better in ID similarity and pose because it encodes the face space into $P$ and $A$ spaces to control the pose and attribute of the face separately.
In addition, MDE achieves higher reconstruction success rates and quality (FID). We attribute the superior performance to the multi-task shared feature extraction network of MDE, making the model have better generalization performance on the original encoding task by sharing the feature representation in the related face perceiving task. Moreover, the multi-task learning strategy can help the DE generate higher-quality reconstructed face images.

To evaluate ALS, we compare it with the fixed latent code swapping module. As can be seen from the quantitative results in Table \ref{ALS}, the swapped images obtained by ALS have more advantages in pose, expression, and image quality. It demonstrates that the proposed ALS helps the swapped image preserve more facial attributes, including but not limited to expression and skin color. While we also note that the ID similarity is slightly inferior to that of the fixed latent code swapping module, which may be because more preserved attribute information from the targets of ALS swapped images has an impact on the ID similarity (see the qualitative results in the supplementary material). 

\begin{table}[!t]
\centering
\begin{tabular}{c|c c c c c}
    \makecell{Method} &
    \makecell{ID simi-\\larity$\uparrow$} & \makecell{Pose$\downarrow$} & \makecell{Expre-\\ssion$\downarrow$} & \makecell{Failure\\Rate$\downarrow$} & \makecell{FID}$\downarrow$\\
    \hline{}
    HieRFE & 0.8725 & 3.81 & \textbf{1.67} & 3.99\% & 47.36  \\
    DE & 0.9076 & 3.20 & 2.12 & 1.73\% & 35.42  \\
    MDE & \textbf{0.9288} & \textbf{2.93} & 2.01 & \textbf{1.73\%} & \textbf{35.17}\\
\end{tabular}
\caption{Quantitative result of the reconstructed faces.}
\label{Ab_Encoder_table1}
\end{table}

\begin{table}[!t]
\centering
\begin{tabular}{c|c c c c}
    \makecell{Method} &
    \makecell{ID simi-\\larity$\uparrow$} & \makecell{Pose$\downarrow$} & \makecell{Expre-\\ssion$\downarrow$} & \makecell{FID}$\downarrow$\\
    \hline{}
    W/o ALS & \textbf{0.4572} & 3.74 & 3.35 & 16.61  \\
    W ALS & 0.4327 & \textbf{3.61} & \textbf{2.70}  & \textbf{15.59}\\
\end{tabular}
\caption{Quantitative result of the effect of ALS.}
\label{ALS}
\vspace{-0.4cm}
\end{table}


\section{Conclusion }
This work presents a novel end-to-end face swapping framework via adaptive latent representation learning. A multi-task dual-space encoder is designed to maintain face details well while perceiving facial regions to generate corresponding masks for blending. We also propose an adaptive latent codes swapping module to avoid too coarse latent codes selection and achieve enhanced attribute preservation. Our experiments demonstrate that our FSALL performs better face swapping than several existing SOTA methods.

{\small
\bibliographystyle{ieee_fullname}
\bibliography{egbib}
}

\end{document}